\documentclass[conference]{IEEEtran}

\usepackage[utf8]{inputenc}
\usepackage[left=19mm, right=19mm, top=25mm, bottom=19mm, paper=letterpaper]{geometry}
\usepackage{graphicx}
\graphicspath{ {./figures/} }

\usepackage{booktabs} 
\newcommand{\xmark}{\ding{55}}
\RequirePackage{amsmath}
\usepackage{amssymb}
\usepackage{pifont}

\newcommand{\norm}[1]{\left\lVert #1 \right\rVert}

\RequirePackage{mathtools}

\RequirePackage{bm}

\usepackage{paralist}

\emergencystretch=\maxdimen
\usepackage{lipsum}

\mathchardef\mhyphen="2D

\usepackage{cite}

\newcommand\blfootnote[1]{%
  \begingroup
  \renewcommand\thefootnote{}\footnote{#1}%
  \addtocounter{footnote}{-1}%
  \endgroup
}

\makeatletter
\def\footnoterule{\kern-3\p@
  \hrule \@width .75in \kern 2.6\p@} 
\makeatother

\usepackage{hyperref}

\begin{document}

\title{Single View Distortion Correction using Semantic Guidance}
\author{
    \IEEEauthorblockN{Szabolcs-Botond Lőrincz$^{1,2}$, Szabolcs Pável$^{1,2}$, Lehel Csató$^{1}$
    }
    \IEEEauthorblockA{$1$ Faculty of Mathematics and Informatics, Babeș-Bolyai University of Cluj-Napoca, Romania\\
    lehel.csato@cs.ubbcluj.ro
    }
    \IEEEauthorblockA{$2$ - Robert Bosch SRL -- Cluj-Napoca, Romania \\
    \{fixed-term.szabolcs.lorincz, szabolcs.pavel\}@ro.bosch.com
    }
}

\maketitle

\begin{abstract}
Most distortion correction methods focus on simple forms of distortion, such as radial or linear distortions. These works undistort images either based on measurements in the presence of a calibration grid \cite{wang2009simple,brauer2000automatic,prescott1997line}, or use multiple views to find point correspondences and predict distortion parameters \cite{stein1997lens,fitzgibbon2001simultaneous,hartley2007parameter}. When possible distortions are more complex, \textit{e.g.} in the case of a camera being placed behind a refractive surface such as glass, the standard method is to use a calibration grid \cite{Dixon2011,wisely2008digital}. Considering a high variety of distortions, it is nonviable to conduct these measurements. In this work, we present a single view distortion correction method which is capable of undistorting images containing arbitrarily complex distortions by exploiting recent advancements in differentiable image sampling introduced by \cite{jaderberg2015spatial} and in the usage of semantic information to augment various tasks. The results of this work show that our model is able to estimate and correct highly complex distortions, and that incorporating semantic information mitigates the process of image undistortion.
\end{abstract}



\section{Introduction}
\label{sec:intro}
\blfootnote{2019 International Joint Conference on Neural Networks (IJCNN)

The final authenticated publication is available online at 

\url{https://doi.org/10.1109/IJCNN.2019.8852065}.}
One way intelligent systems are able to perceive and interact with complex environments is vision, thus, they are highly reliant on a wide variety of computer vision algorithms, such as object detection, semantic segmentation, or depth estimation. The propagation of errors caused by geometric image distortions has a negative effect on the accuracy of these algorithms, therefore it is critical to correct them.

Camera based driver assistance and autonomous driving systems are no exception, as the camera is usually placed behind the vehicle's windshield, which typically consists of two curved sheets of glass with a plastic layer laminated between them. The curvature, deviation in thickness and inconsistency in the parallelism of the two surfaces, causes geometric distortions. Measuring ground truth distortions caused by various glass surfaces requires laboratory setups, making the collection of large training sets – and as a consequence using standard supervised learning – unfeasible.

Our contribution is threefold. First, we present a scalable deep learning approach that can correct arbitrarily complex nonlinear distortions. Second, we construct two data sets comprising of real-world (KITTI odometry \cite{Geiger2012CVPR}) and synthesized (Carla \cite{Dosovitskiy17}) images and corresponding semantic segmentation, on which we apply parametric distortions sampled from a distribution derived from real-world measurements in the presence of different windshields. Third, we train our network in an end-to-end manner without using hard to obtain ground truth distortions as supervision, and instead leverage recent advancements in differentiable image sampling to formulate a loss based on Multi-Scale Structural Similarity Index Metric (MS-SSIM) \cite{wang2003multiscale}.

Our experiments on both data sets show that our model is able to estimate highly complex distortions. Moreover, the network does not only estimate the distortions, but it produces directly the undistorted image and segmentation also.

\begin{figure}[t]
	\centering
		\centerline{\includegraphics[width=\columnwidth]{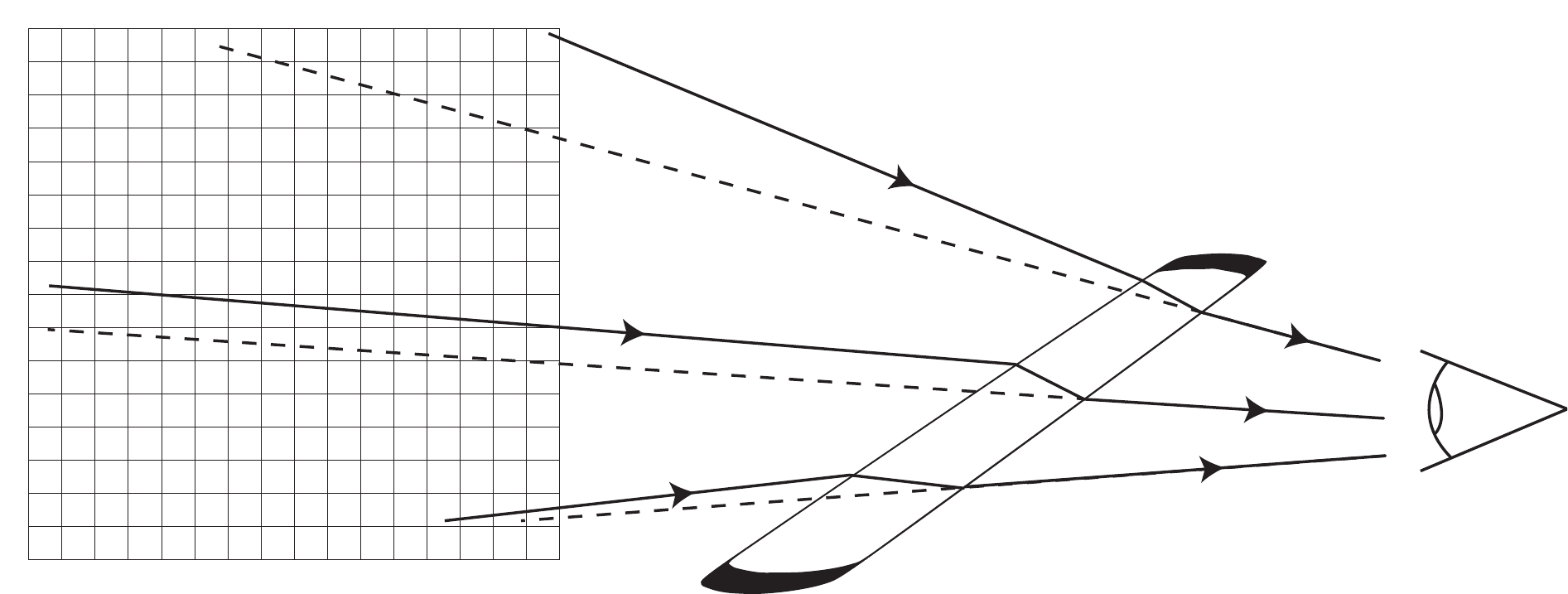}}
		\caption{Computer vision pipelines in autonomous driving systems are highly affected by geometric distortions, due to the camera being placed behind the vehicle's windshield in most cases. These are often curved, have variable thickness, and are tilted, resulting in complex, nonlinear distortions.
		}
		\label{fig:tps}
\end{figure}

\section{Previous Work}
\label{sec:lit}
In this section, we summarize the studies related to our approach. First, a brief overview of existing work addressing the problem of geometric distortion correction is provided. Second, various successful use cases of semantic guidance are presented, including distortion correction.
Third, current progress of spatial transformer networks and their applications is described, which is a main unit of our distortion correction system.

\subsection{Distortion Correction}
\label{sec:distcorr}
Most of the literature focuses on simple forms of distortions, such as radial distortions. According to \cite{zhang2000flexible},
\newgeometry{left=19mm, right=19mm, top=19mm, bottom=19mm}
\noindent radial distortions are the main components of distortions caused by lenses in addition to decentering distortions and thin prism distortions. There are two major approaches for correcting radial distortions in the literature. The first involves using point correspondences of two or more images \cite{stein1997lens,fitzgibbon2001simultaneous,hartley2007parameter}, while the second is based on finding  distorted straight lines in single images and estimating the distortion parameters \cite{wang2009simple,brauer2000automatic,prescott1997line}.

In contrast with these methods, in \cite{10.1007/978-3-319-54187-7_3} a CNN based model is introduced, which predicts radial distortion parameters based on single input images. In \cite{yin2018fisheyerecnet} fisheye distortion parameters are estimated by CNNs. Another achievement is that both networks in \cite{10.1007/978-3-319-54187-7_3} and \cite{yin2018fisheyerecnet} are trained on synthetically distorted images, but it is demonstrated that they achieve similarly good results in undistorting images containing real distortions.

Only a few studies have conducted experiments for more complex distortions. In \cite{Dixon2011} a calibration grid is used to measure aircraft windscreen distortions and a decision-tree-based classifier is introduced which classifies the distortions as acceptable or not. In \cite{sato2006visual} and \cite{wientapper2013camera}, the distortions caused by a car's front window are estimated with the help of a calibration grid in order to create head-up displays (HUD), which show important information directly in the field of view of the driver. The disadvantage of utilizing a calibration grid is that the measurements need to be conducted for each kind of distortion separately -- hence the method is not scalable.

\subsection{Semantic Guidance}
\label{sec:semguid}
Semantic segmentation is understanding an image at pixel level i.e, an object class (\textit{e.g.} car, road, pedestrian, ...) is assigned to each pixel in a given image. A number of studies have investigated the potential utilization of semantic labels for solving various problems.

In \cite{tsai2017deep}, an end-to-end deep convolutional neural network is proposed, which learns to capture semantic information, and uses that information for image harmonization. A multi-context embedding network, which integrates high-level semantic labels and low-level image details is proposed for automatic shadow removal from single images in \cite{qu2017deshadownet}. Single image depth estimation is achieved in \cite{liu2010single} by first performing a semantic segmentation of the scene and using the semantic labels to guide the 3D reconstruction. The task of optical flow estimation is also enhanced by modelling motions as a function of the classes present in the images \cite{sevilla2016optical}.

By knowing semantic labels, the correction of distortions can also be augmented, due to objects of different classes having different geometric properties (\textit{e.g.} buildings have straight borders, while cars usually do not). This property is exploited in \cite{yin2018fisheyerecnet} to facilitate estimating fisheye distortion parameters and correcting the distortions.

\subsection{Spatial Transformer Module}
\label{sec:stn}
Spatial transformers introduced in \cite{jaderberg2015spatial} are modules which can be incorporated into any CNN to augment various problems, and are composed of three parts. The first part is the localization network, which takes an input feature map or input image and produces the parameters of the chosen transformation. The second generates a sampling grid based on the predicted transformation parameters. The third part is a differentiable sampler, which transforms the input using the generated sampling grid. The task of these modules is to achieve real spatial invariance by automatically transforming input images or feature maps to a prototype instance before they are used for classification or other tasks. Recent successful use-cases of spatial transformers include handwritten digit classification \cite{jaderberg2015spatial} on distorted MNIST \cite{lecun2010mnist} data set, recognition of sequences of numbers \cite{jaderberg2015spatial} on Street View House Numbers (SVHN) \cite{netzer2011reading} and scene text recognition \cite{shi2016robust}.

\section{Data sets}
\label{sec:data}
In order to demonstrate that our model is able to undistort both synthetic and real-world images, we construct two data sets, Distorted Carla (DC) and Distorted KITTI (DK).

Distorted Carla is composed of $10,000$ synthetic images and their corresponding semantic labels generated using Carla driving simulator \cite{Dosovitskiy17}. The labels which are provided by Carla are presented in Table \ref{tab:semlabels}. We generate the images with autopilot turned on, at a fixed time-step of 0.2 seconds, using weather preset ClearNoon, having 128 vehicles and 256 pedestrians spawned on the map Town01. 

Distorted KITTI is comprised of $15,223$ images originating from sequences $00$ to $06$ of KITTI odometry \cite{Geiger2012CVPR} data set.

\begin{table}
	\caption{Semantic labels provided by Carla}
	\label{tab:semlabels}
	    \begin{center}
    		\begin{small}
    			\begin{sc}
    				\begin{tabular}{lc}
    					\toprule
    					Value   & Label       \\
    				    \midrule
    					0       & None      \\
                        1       & Buildings \\
                        2       & Fences    \\
                        3       & Other     \\
                        4       & Pedestrians \\
                        5       & Poles     \\
                        6       & Road lines \\
                        7       & Roads     \\
                        8       & Sidewalks \\ 
                        9       & Vegetation \\
                        10      & Vehicles  \\
                        11      & Walls     \\
                        12      & Traffic signs \\
    					\bottomrule
    				\end{tabular}
    			\end{sc}
    		\end{small}
        \end{center}
\end{table}

The geometric distortions are defined as a grid containing displacement vectors $\bm{\delta}=\lbrack \delta_{x_{i}},\delta_{y_{i}}\rbrack^{\top}$ for each pixel in the image. We apply distortions synthetically on the images and corresponding semantic labels. As a second step, we interpolate the color values in the sampled images using bilinear interpolation, and the semantic labels using nearest-neighbour interpolation. The set of distortions used in our experiments is drawn from real-world windscreen distortion distribution data, covering a wide range of parameter settings.


\section{Our approach}

First, we introduce our parametric distortion model. Then, we specify the architecture of our deep network. We close this section with a description of how our model is trained.

\subsection{Distortion Model}
\label{sec:distmodel}

In \cite{bookstein1989principal} it is shown, that a pair of thin plate splines (TPS), one representing the $x$-component and the other the $y$-component form a map from $\mathbb{R}^{2}$ to $\mathbb{R}^{2}$, which can model biological deformations, \textit{e.g.} in the case of Apert syndrome. We model geometric distortions with thin plate spline pairs.

The transformed coordinates $\bm{f}_{tps}(G_{i})$ at image coordinate $G_{i}=\lbrack x_{i},y_{i} \rbrack ^{\top}$ assuming $n$ control points are given, are defined as
\begin{equation}
    \bm{f}_{tps}(G_{i}) = A
    \left[\begin{matrix}G_{i} \\ 1 \end{matrix}\right] + \sum_{k=1}^{n} \varphi(\norm{\bm{p}^{\prime}_{k}-G_{i}}_2) \cdot \bm{w}_k.
    \label{eqn:tps}
\end{equation}

For our purpose, we use $n=16$ control points, but it is possible to use more points to model arbitrarily complex distortions. The target control points $P^{\prime}=\lbrack\bm{p}^{\prime}_1,\bm{p}^{\prime}_2,\dots,\bm{p}^{\prime}_{n}\rbrack\in \mathbb{R}^{2\times n}$ in our case are fixed and evenly distributed on a $4\times4$ grid, whereas source control points $P=\lbrack\bm{p}_1,\bm{p}_2,\dots,\bm{p}_{n}\rbrack \in \mathbb{R}^{2\times n}$ have to be localized on the distorted image in order to interpolate the displacements between them.

The first term of $\bm{f}_{tps}(\cdot)$ is an affine transformation $A=\lbrack \bm{a_1},\bm{a_2},\bm{a_3}\rbrack \in \mathbb{R}^{2 \times 3}$. The second term represents the non-affine deformation, where the radial basis kernel corresponding to TPS transformation is $\varphi(r) = r^{2}log(r)$, with $r$ denoting the Euclidean distance between two points, while $W=\lbrack\bm{w}_1,\bm{w}_2,\dots,\bm{w}_n\rbrack \in \mathbb{R}^{2\times n}$ is a warping coefficient matrix.
The transformation parameter $\theta$ containing the two terms is calculated by
\begin{equation}
    \theta=(W\vert A)^{\top}=L^{-1}
    \begin{bmatrix}
        P^{\top} \\
        \bm{0}^{3\times 2}
    \end{bmatrix},
    \label{eqn:transf}
\end{equation}
where $L^{-1}$ is the inverse of the padded kernel matrix $L$ which is computed based on the target control points and is given by
\begin{equation}
    L=
    \begin{bmatrix}
        K & \bm{1}^{n\times1} & P^{\prime\top} \\
        \bm{1}^{1\times n} & 0 & \bm{0}^{1 \times 2} \\
        P^{\prime} & \bm{0}^{2 \times 1} & \bm{0}^{2 \times 2}
    \end{bmatrix}.
    \label{eqn:pkern}
\end{equation}
Here, $K\in \mathbb{R}^{n\times n}$ is defined by
\begin{equation}
    K = 
    \begin{bmatrix}
        0 & \varphi(r_{12}) & \cdots & \varphi(r_{1n})\\
        \varphi(r_{21}) & 0 & \cdots & \varphi(r_{2n})\\
        \vdots & \vdots     & \ddots & \vdots         \\
        \varphi(r_{n1}) & \varphi(r_{n2}) & \cdots & 0
    \end{bmatrix},
    \label{eqn:kern}
\end{equation}
where $r_{ij}$ denotes the Euclidean distance between target control points $\bm{p}^{\prime}_i$ and $\bm{p}^{\prime}_j$.

In order to undistort images we need the displaced coordinates for each point in the undistorted reference grid $G=\lbrack G_{1}, G_{2}, \cdots, G_{N} \rbrack$, where $N$ is the total number of pixels in the undistorted image. Thus, we first define matrix $K^{\prime} \in \mathbb{R}^{N \times n}$ which contains radial basis kernel values of the pairwise distances of undistorted grid points and target control points and is defined by
\begin{equation}
    K^{\prime} = 
    \begin{bmatrix}
        \varphi(r_{11}^{\prime}) & \varphi(r_{12}^{\prime}) & \cdots & \varphi(r_{1n}^{\prime})\\
        \varphi(r_{21}^{\prime}) & \varphi(r_{22}^{\prime}) & \cdots & \varphi(r_{2n}^{\prime})\\
        \vdots & \vdots     & \ddots & \vdots         \\
        \varphi(r_{N, 1}^{\prime}) & \varphi(r_{N, 2}^{\prime}) & \cdots & \varphi(r_{N, n}^{\prime})
    \end{bmatrix},
    \label{eqn:kern2}
\end{equation}
where $r_{ij}^{\prime}$ denotes the Euclidean distance between grid point $G_{i}$ and  target control point $\bm{p}^{\prime}_j$.

Let $\tau_\theta(G)$ be the distorted grid consisting of displaced coordinates, where $\tau_{\theta}$ is a transformation of choice parameterized by $\theta$. The distorted grid is given by
\begin{equation}
    \tau_{\theta}(G) = \begin{bmatrix} K^{\prime} & \bm{1}^{N \times 1} & G^{\top} \end{bmatrix} \theta.
    \label{eqn:src}
\end{equation}

\begin{figure}[t]
	\centering
		\centerline{\includegraphics[width=\columnwidth]{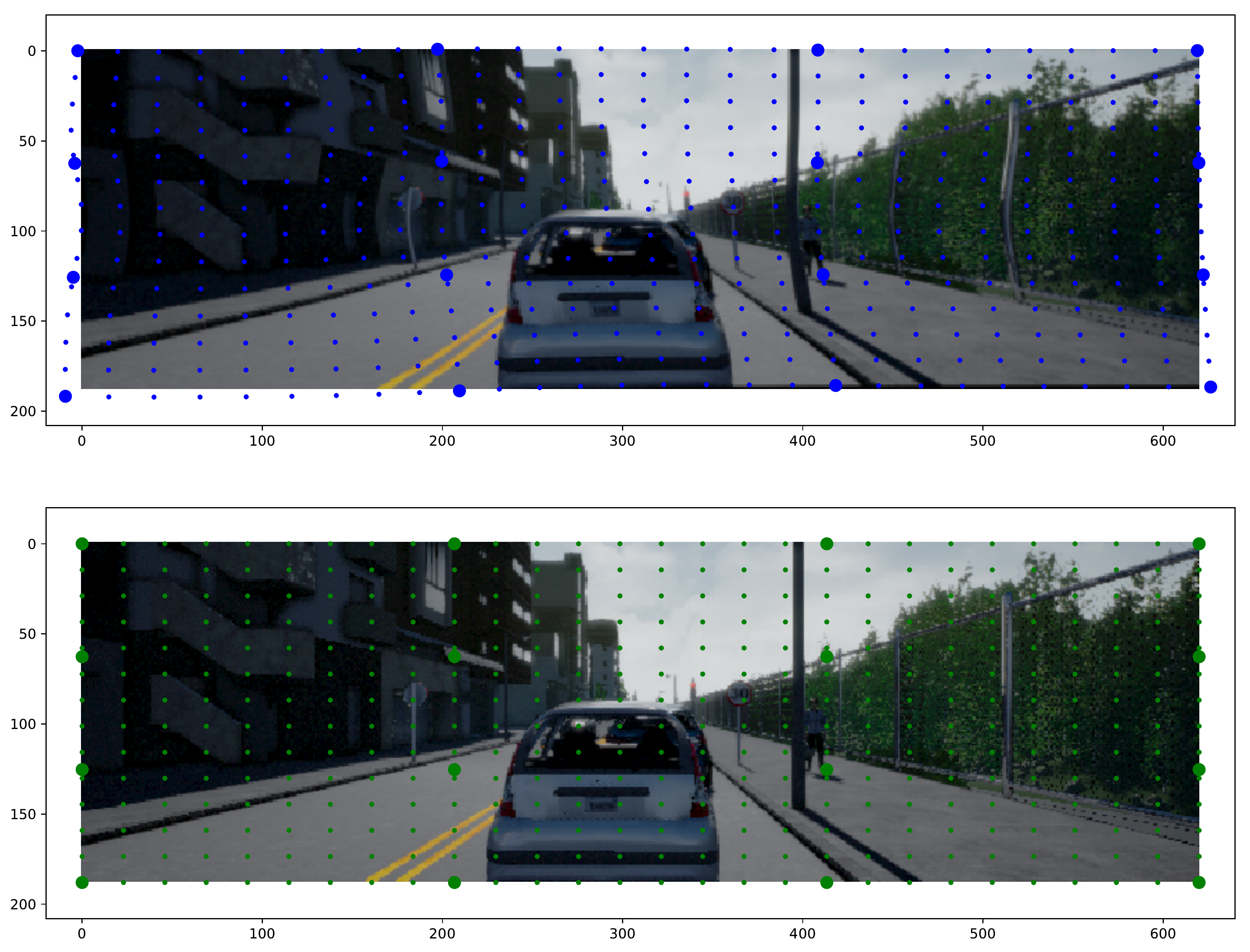}}
		\caption{Sample distorted image (top) and undistorted image (bottom) from Distorted Carla data set. The task is to localize source control points in the distorted image (large blue dots). Since the target control points (large green dots) are fixed, we are able to generate a sampling grid (small blue dots) using thin plate spline interpolation, based on which we can undistort the image.}
		\label{fig:dataset}
\end{figure}

\subsection{Proposed Architecture}
\label{sec:arch}

\begin{figure}[t!]
    \centering
		\centerline{\includegraphics[width=\columnwidth]{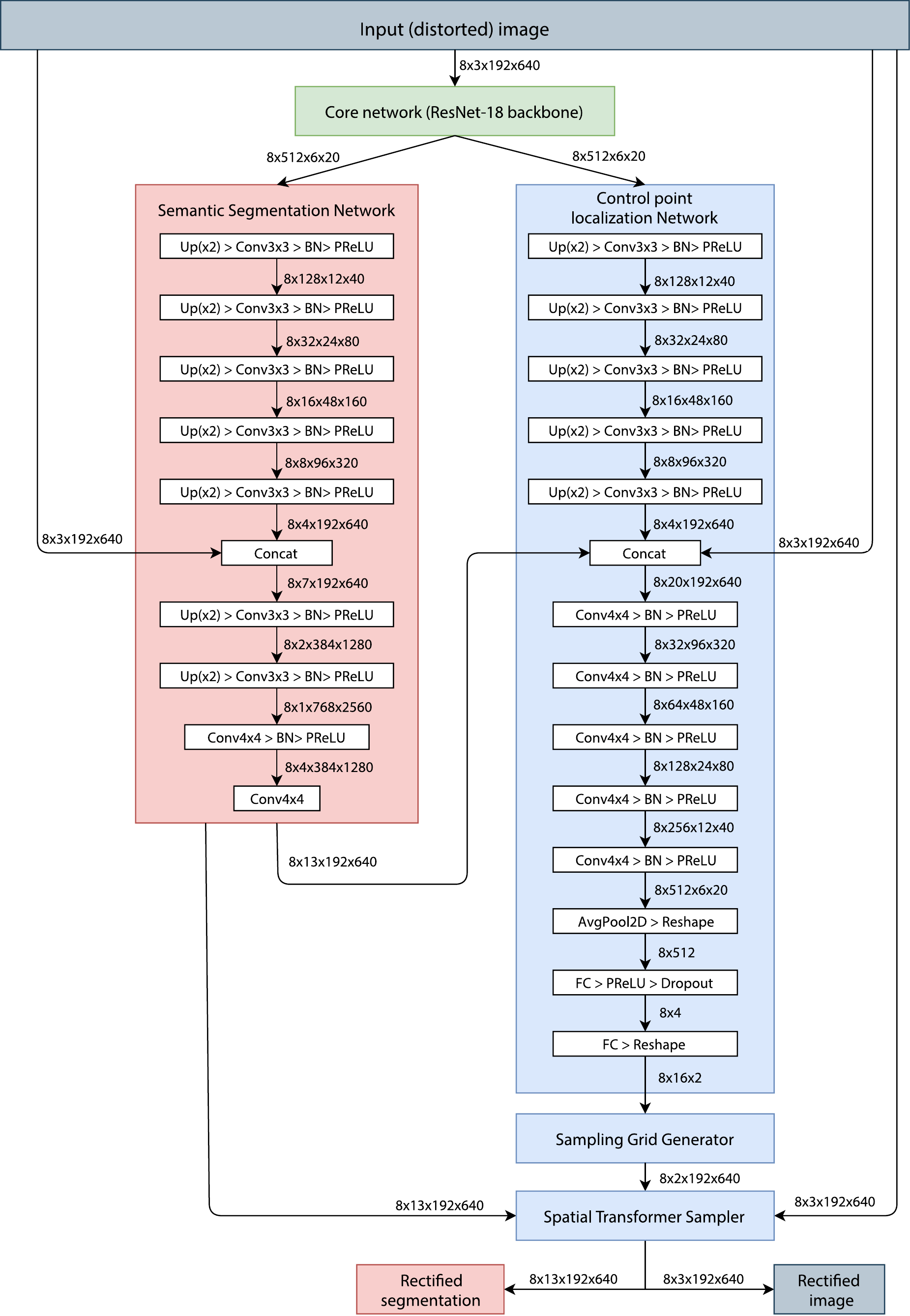}}
		\caption{Schematic diagram of the proposed model composed of three main processing units: the core network (green), the semantic segmentation network (red), and the spatial transformer module (light blue). First, low-level features are extracted using the core network. The semantic segmentation module then combines the input image with the upsampled low-level features and produces pixel-level semantic labels. The spatial transformer module fuses the extracted low-level features with high-level semantics and the input image, and predicts the transformation parameters, based on which the sampling grid is generated, and the output (undistorted) image and segmentation are produced.
	    }
		\label{fig:arch}
\end{figure}

We employ an end-to-end architecture which takes a single distorted image $I$ as input, and outputs the undistorted image $I^{\prime}$ and its corresponding semantic labels (Figure \ref{fig:arch}). Our architecture processes the input in two steps: a feature extraction step, and a distortion correction step. \\


\noindent \textbf{Feature Extraction}: First, low-level features are extracted by the core network, for which we use ResNet-18 \cite{He_2016_CVPR} pre-trained on ImageNet \cite{Russakovsky2015} and remove the top two layers.

The model includes a semantic segmentation network, as in \cite{yin2018fisheyerecnet}, which provides high-level semantics for undistorting the images. The segmentation network takes the extracted feature maps with channel dimension $512$ and the distorted image as input and outputs high-level semantics. The feature maps are first upsampled using five resize-convolution layers \cite{odena2016deconvolution}. Each of these layers upsamples the input feature map by a factor of two using nearest-neighbour sampling (Upx2), then applies a convolution with kernel size $3$, stride $1$ and padding $1$, followed by batch normalization \cite{Ioffe_2015_ICML} and a parametric rectified linear unit (PReLU) \cite{he2015delving}.

After five such blocks, the upsampled feature maps are concatenated with the input image, and are further upsampled using two resize-convolution layers. They are then passed to a Conv4x4-BN-PReLU block. Finally, a Conv4x4 layer produces the pixel-level semantic labels, corresponding to the $13$ classes which Carla provides as ground truth.\newline

\begin{figure}[t!]
    \centering
		\centerline{\includegraphics[width=\columnwidth]{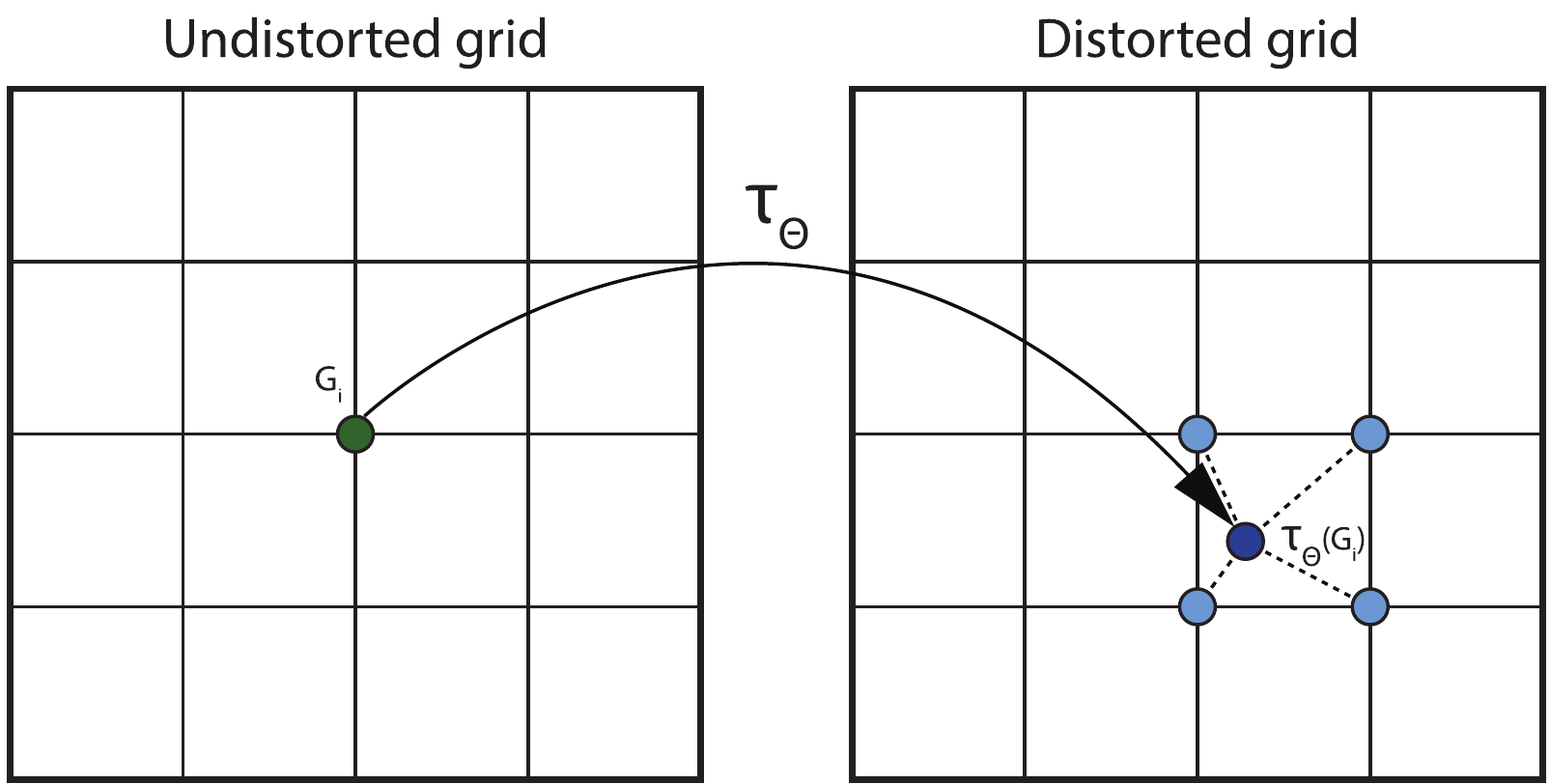}}
		\caption{The undistorted grid points $G_i$ (green) are transformed to distorted grid points $\tau_{\theta}(G_i)$ (dark blue). The pixel values in the undistorted image are calculated by bilinear sampling from the nearby pixel values (light blue). The undistorted semantic labels are the semantic labels nearest to the transformed points.
	    }
		\label{fig:sampl}
\end{figure}

\noindent \textbf{Distortion Correction}:
Once the control points have been localized, the model generates a sampling grid containing 2D pixel coordinates using Equation \eqref{eqn:src}. Finally, a Spatial Transformer Sampler \cite{jaderberg2015spatial} takes the sampling grid, the distorted image and semantic segmentation to produce the undistorted image and segmentation (Figure \ref{fig:sampl}). Both the sampling grid generator and the sampler are differentiable, thus, end-to-end learning using gradient descent is possible.

\subsection{Model Training}
\label{sec:train}
In the section that follows, we detail our training method, including network initialization, training loss function and hyperparameters.\\

\noindent{\textbf{Network Initialization}:} The model parameters are initialized using "He" uniform initialization described in \cite{he2015delving}, except for the last fully connected layer in the control point localization network, where weights are set to zero, and biases are set to initially produce the target control points, similarly to \cite{shi2016robust}.\\

\noindent{\textbf{Training Loss}:} We experiment with various settings of training loss functions. In order to enforce reconstruction of both the image and the semantic labels, we employ a joint loss.

For the former, we propose to use the reconstruction loss $ \mathcal{L}_{r}$ based on MS-SSIM \cite{wang2003multiscale}, which involves computing single-scale structural similarity on multiple scales, measuring the similarity of two image patches in luminance, contrast, and structure. This metric is proper in case of training our model on data sets containing real-world distortions, since ground truth sampling grid is hard to obtain in those scenarios. The proposed loss is given by

\begin{equation}
    \mathcal{L}_{r}=-\frac{\textup{MS-SSIM}(I, I^{\prime})+1}{2}.
    \label{eqn:Lmsssim}
\end{equation}

Since we applied synthetic distortions, we are able to calculate the ground truth sampling grid for undistorting the images. Therefore, we investigate whether minimizing directly the grid loss gives better results than $ \mathcal{L}_{r}$ from Equation \eqref{eqn:Lmsssim}. The grid loss is formulated as

\begin{equation}
    \mathcal{L}_g= \frac{1}{N} \sum_{i=1}^{N}{\norm{\tau_{\theta}(G_i)- \hat{\tau}_{\theta}(G_i)}}^2_2,
    \label{eqn:LGrid}
\end{equation}
where $\tau_{\theta}(G)$ is the estimated sampling grid, while $\hat{\tau}_{\theta}(G)$ is the ground truth sampling grid.

For semantic segmentation, we use $\mathcal{L}_{s}$, which we calculate as the mean of pixel-wise cross-entropy between the ground truth distorted and predicted distorted semantic labels.

The final loss of our network is composed of a weighted sum of all three losses:

\begin{equation}
    \mathcal{L}= \mathcal{L}_r + \lambda_1 \mathcal{L}_g + \lambda_2 \mathcal{L}_s,
    \label{eqn:loss}
\end{equation}
where $\{\lambda_i\}$ is a set of weighting coefficients to balance the loss functions. We set $\lambda_1=100$, $\lambda_2=0.25$ empirically.\\

\noindent{\textbf{Training Method}:} We use Adam \cite{kingma2014adam} optimization and mini-batch gradient-descent, with batch size of $8$. The learning rate is set to $10^{-3}$, except for the core network, where we use $5 \cdot 10^{-4}$. During fine-tuning on sequences $01$ to $06$ of Distorted KITTI, we set the learning rate of the semantic segmentation network to $0$, because no ground truth semantic labels are provided.

\section{Experimental Results}
\label{sec:res}

We split Distorted Carla and Distorted KITTI into training and test sets. We train the networks on Distorted Carla Train (8,000 images) for 10 epochs, while we test the models on both Distorted Carla Test (2,000 images) and on sequence $00$ of Distorted KITTI (4,539 images). We fine-tune the previously trained networks on sequence $01-06$ of Distorted KITTI (10,684 images) for another 10 epochs and test the fine-tuned models on sequence $00$.

Since we synthetically generate the applied distortions, we are able to measure the performance of our model quantitatively by calculating the residual distortion norm measured in pixels for each pixel on each image and computing the mean and standard deviation over all the pixels in all images, contrary to previous work \cite{yin2018fisheyerecnet} where only qualitative metrics were used, such as Peak Signal-to-Noise Ratio (PSNR) or MS-SSIM \cite{wang2003multiscale}.

In Table \ref{tab:orig} we report mean and standard deviation of distortion norms in pixels in the original distorted test images, which we use as a comparison to the residual distortion error norms after distortion correction. The mean and standard deviation of residual distortion norms in pixels after distortion correction for different settings of our method are summarized in Table \ref{tab:results}.

\begin{table}
	\caption{Original distortion norm (px)}
	\label{tab:orig}
	    \begin{center}
    		\begin{small}
    			\begin{sc}
    				\begin{tabular}{lcc}
    					\toprule
    					Data set & Mean & Std \\
    					\midrule
    					Distorted Carla Test & 8.46 & 3.92   \\
    					Distorted KITTI Test & 8.59 & 3.32   \\
    					\bottomrule
    				\end{tabular}
    			\end{sc}
    		\end{small}
        \end{center}
\end{table}

\begin{table}[t]
\centering
    \caption{Residual distortion norm (px)}
	\label{tab:results}
	\begin{small}
		\begin{sc}
			\begin{tabular}{lcccccc}
			    \toprule
			    Fine-tune & Test & $\mathcal{L}_r$ & $\mathcal{L}_g$ & $\mathcal{L}_s$ & Mean & Std \\
			    \midrule
 		        \xmark & DC Test & \checkmark & & & 2.26 & 1.49 \\
 		         & & & \checkmark & & 2.25 & 1.59 \\
 		         & & \checkmark & \checkmark & & 2.15 & 1.47 \\
 		         & & \checkmark & & \checkmark & $\bm{1.98}$ & $\bm{1.40}$ \\
 		         & & & \checkmark & \checkmark & 2.15 & 1.53 \\
 		         & & \checkmark & \checkmark & \checkmark & 2.06 & 1.45 \\
   		         \midrule
 		         \xmark & DK 00 & \checkmark & & & 1.75 & 1.10 \\
 		         & & & \checkmark & & 2.28 & 1.52 \\
 		         & & \checkmark & \checkmark & & 1.99 & 1.35 \\
 		         & & \checkmark & & \checkmark & 1.65 & 1.09 \\
 		         & & & \checkmark & \checkmark & $\bm{1.37}$ & $\bm{0.88}$ \\
 		         & & \checkmark & \checkmark & \checkmark & 2.53 & 1.31 \\
 		         \midrule
 		         DK 01-06 & DK 00 & \checkmark & & & 1.24 & 0.72 \\
 		         & & & \checkmark & & 1.33 & 0.67 \\
 		         & & \checkmark & \checkmark & & 1.30 & 0.70 \\
 		         & & \checkmark & & \checkmark & 1.30 & 0.69 \\
 		         & & & \checkmark & \checkmark & $\bm{1.22}$ & $\bm{0.72}$ \\
 		         & & \checkmark & \checkmark & \checkmark & 1.35 & 0.72 \\
			    \bottomrule
			\end{tabular}
		\end{sc}
	\end{small}
\end{table}

We conducted various experiments using different configurations of the training loss function. It can be seen, that employing semantic loss reduces the residual error in general, except in the case where it is used alongside both reconstruction and grid loss, when tested on Distorted KITTI.

The model which performs the best on Distorted Carla Test uses reconstruction loss and segmentation loss, whereas on Distorted KITTI $00$, the best performing model uses grid loss together with segmentation loss. Among the fine-tuned models, the model which achieves the lowest mean residual distortion norm is also the one using grid loss and segmentation loss.

The fine-tuned models do not achieve significantly better performance when using grid loss instead of reconstruction loss, so it is possible to train our network without obtaining ground truth sampling grid, which simplifies the training and usage of our model.

\section{Conclusion and future work}
\label{sec:conc}

In this work we have demonstrated a deep-network based system that can correct arbitrarily complex distortions and have illustrated its accuracy for the highly nonlinear distortions caused by vehicles' windshields. We have also shown, that training on synthesized data, the model is able to generalize to real-world scenes. Additionally, we demonstrated that incorporating semantic information mitigates the process of image undistortion. It is possible to train our model using real-world distortions instead of synthetic distortions, since only the distorted image is needed as input, and the undistorted image as supervision for the fine-tuning phase. In the future, we want to test the model in real-world scenarios, and to quantify the impact of our algorithm on the performance of end-to-end computer vision pipelines.

\section*{Acknowledgment}

The authors thank Robert Bosch SRL for technical support and for useful discussions.

\bibliographystyle{IEEEtran}
\bibliography{dist}
\end{document}